\documentclass[runningheads]{llncs}
\usepackage[T1]{fontenc}

\usepackage{graphicx,verbatim}
\usepackage{comment}
\usepackage{bbm}
\usepackage{bbold}
\usepackage{tabularx}
\usepackage{marvosym}
\newcolumntype{C}{>{\centering\arraybackslash}X}

\usepackage{xcolor}
\usepackage{amsmath}

\begin{document}
\title{3D Classification of Paramagnetic Rim Lesions in Multiple Sclerosis via Asymmetric QSM–FLAIR Modeling}

\titlerunning{Classification of paramagnetic rim lesions}

\author{Veronica Pignedoli\inst{1,\mbox{\Letter}}, Giacomo Boffa\inst{2,3}, Nicoletta Noceti\inst{1}, Matilde Inglese\inst{2,3}, Francesca Odone\inst{1}, Matteo Moro\inst{1}}  
\authorrunning{V. Pignedoli et al.}
\institute{MaLGa, DIBRIS, University of Genova, Genova, Italy \and DINOGMI, University of Genova, Genova, Italy \and IRCCS Azienda Ospedaliera Metropolitana, Genova, Italy  \\
   \mbox{\Letter} \email{veronica.pignedoli@edu.unige.it}}
  
\maketitle              
\begin{abstract}

Paramagnetic rim lesions (Rim$^+$) identified on susceptibility-sensitive MRI have recently emerged as a specific biomarker of chronic active inflammation in Multiple Sclerosis (MS) and are associated with long-term disability progression. However, susceptibility imaging and expert interpretation remain limited to specialized centers, visual assessment is time-consuming and variable, and the low prevalence of Rim$^+$ lesions poses severe class imbalance challenges for automated analysis.
We propose a 3D multimodal deep learning framework for lesion-level Rim$^+$/Rim$^-$ classification from Quantitative Susceptibility Mapping (QSM) and FLAIR MRI. The architecture explicitly models modality asymmetry by treating QSM as the primary susceptibility-driven signal and conditioning it with FLAIR-derived structural context. To improve robustness under limited data, we employ self-supervised multimodal pretraining followed by supervised fine-tuning with contrastive regularization.
The method was evaluated on a clinically acquired cohort of 88 people with MS with expert lesion annotations as reference standard. Results highlight improved performance compared to prior architectures, supporting the effectiveness of asymmetric multimodal modeling for automated chronic active lesion identification.


\end{abstract}

\section{Introduction} 
\textit{Clinical problem formulation.}
Multiple sclerosis (MS) is a chronic inflammatory disease of the central nervous system characterized by focal demyelinating lesions with heterogeneous pathological evolution. Among these, chronic active lesions (Rim$^+$), also referred to as paramagnetic rim lesions (PRL), are defined by a rim of iron-laden microglia and macrophages surrounding a demyelinated core and are associated with increased axonal damage and worse long-term disability progression \cite{kuhlmann2017updated,absinta2019association,dal2024chronic}. In this context, lesions without this feature are referred to as Rim$^-$. The clinical relevance of Rim$^+$ lesions has recently been reinforced by the 2024 revision of the McDonald diagnostic criteria, which recognizes advanced imaging biomarkers as supportive evidence to improve diagnostic specificity \cite{montalban2025diagnosis}. In routine clinical practice, structural MRI sequences such as Fluid Attenuated Inversion Recovery (FLAIR) and T1-weighted (T1w) imaging are commonly acquired and are sensitive to lesion detection, but provide limited specificity regarding the underlying tissue substrate. Instead, Quantitative Susceptibility Mapping (QSM), a quantitative MRI technique that estimates magnetic susceptibility ($\chi$), allows for in vivo assessment of iron accumulation and has emerged as a key modality for the identification of Rim$^+$ lesions \cite{tranfa2022quantitative}. 
However, visual Rim$^+$ identification in QSM is time-consuming, expertise-dependent, and susceptible to confounding factors such as veins and imaging artifacts \cite{elliott2023lesion}. These limitations call for automated  classification procedures.
 In this work, we address the problem of automatic lesion-level (Rim$^+$/Rim$^-$) classification from lesion-centered MRI patches extracted from co-registered QSM and FLAIR volumes.
 It is worth mentioning that this task is challenged by severe class imbalance, as Rim$^+$ lesions represent a small fraction of total lesions \cite{rahmanzadeh2022new}, and by the absence of publicly available benchmark datasets. 

\vspace{0.1cm}
\noindent \textit{State of the art.}
Early computational approaches relied on susceptibility phase imaging. Barquero et al. \cite{barquero2020rimnet} proposed RimNet, a 3D patch-based convolutional neural network operating on phase and FLAIR images through parallel VGG-inspired branches. Lou et al. \cite{lou2021fully} introduced the Automated Paramagnetic Rim Lesion (APRL) classifier, which leverages handcrafted radiomic features combined with SMOTE-based imbalance handling \cite{chawla2002smote}. More recently, Zhang et al. \cite{zhang2022qsmrim} proposed QSMRim-Net, an imbalance-aware architecture that integrates convolutional features from QSM and FLAIR with a large set of lesion-level radiomic descriptors derived from QSM, coupled with DeepSMOTE-based feature synthesis. QSMRim-Net represents the most comprehensive QSM-based framework currently available. However, its reliance on handcrafted radiomic features and latent-space oversampling introduces additional computational complexity and may limit generalization across heterogeneous clinical datasets. Motivated by these considerations and by preliminary experiments on our clinical cohort, we explore an alternative end-to-end strategy.

\vspace{0.1cm}
\noindent \textit{Our contributions.}
We propose an end-to-end multimodal framework for Rim$^+$/ Rim$^-$ lesion classification operating directly on lesion-centered QSM and FLAIR patches. The architecture explicitly models modality asymmetry: QSM encodes the primary susceptibility-driven signature, while FLAIR provides structural context through spatial feature-wise linear modulation (FiLM) \cite{perez2018film}. To improve robustness under extreme class imbalance and limited sample size, we incorporate self-supervised cross-modal pretraining and supervised contrastive regularization. The proposed method is evaluated on a cohort of 88 people with MS (pwMS) acquired in a routine clinical setting, reflecting heterogeneous imaging conditions and real-world Rim$^+$ prevalence. Our main contributions are: (i) an end-to-end multimodal lesion-level classifier based on QSM and FLAIR (\url{https://github.com/veronicapignedoli/FRODO}); (ii) an asymmetric conditioning strategy that preserves susceptibility-specific discriminative features while integrating structural context; and (iii) a comprehensive evaluation on a clinically representative dataset with statistical comparison against prior approaches.

\section{Method}
\begin{figure}
\centering
\includegraphics[width=\textwidth]{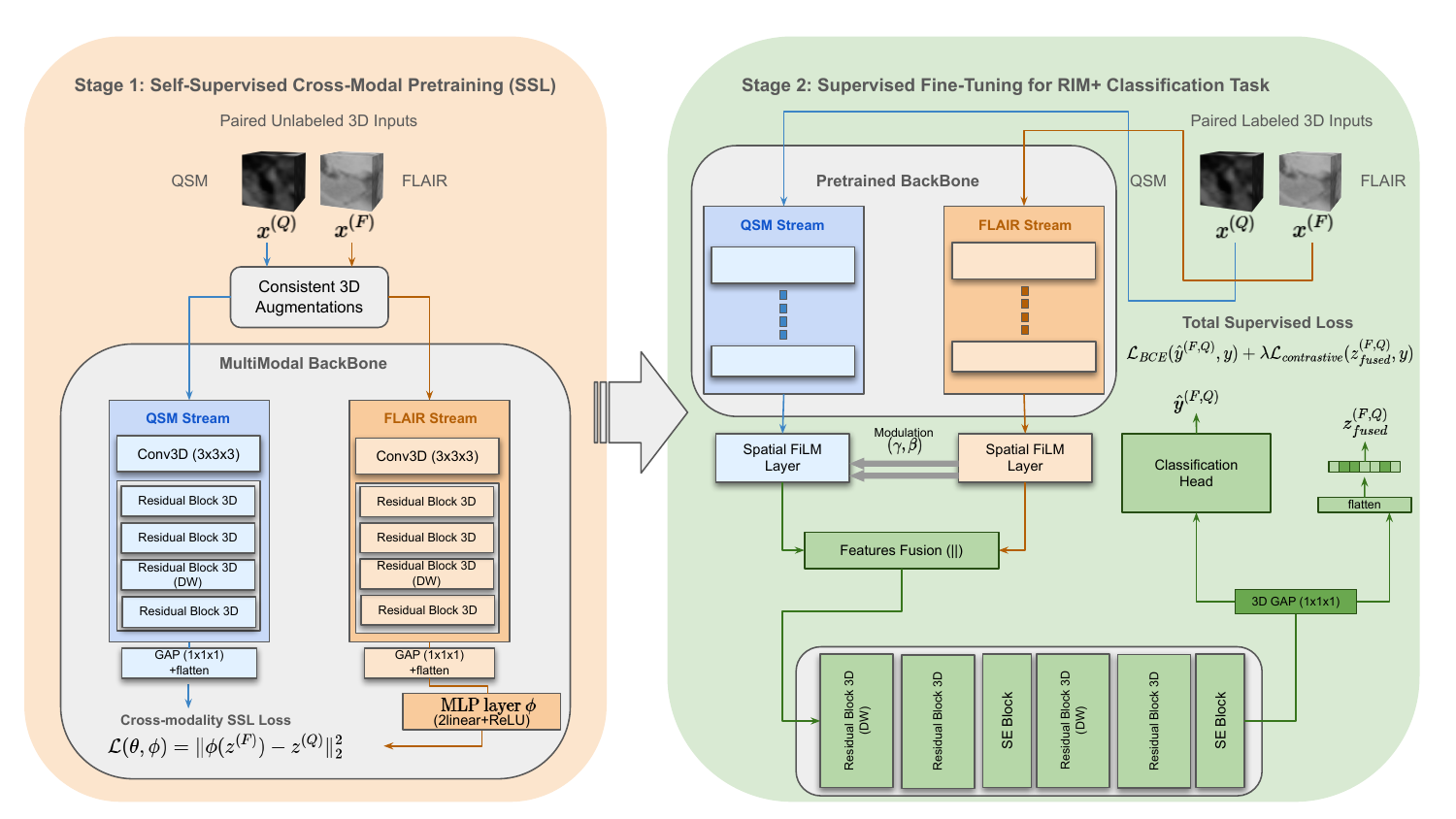}
\caption{Overview of the proposed multi-modal 3D classification framework for Rim$^+$/Rim$^-$ prediction. 
The network is first pretrained using a self-supervised contrastive objective and then fine-tuned with supervised binary classification loss.} \label{fig:architecture}
\end{figure}

 Our method explicitly models the asymmetric physiological roles of the two MRI modalities: QSM encodes the primary susceptibility-driven rim signature, while FLAIR provides structural context through spatial feature-wise linear modulation (FiLM) \cite{perez2018film}. An overview is shown in Figure \ref{fig:architecture}. The method follows a two-stage training strategy: first, self-supervised pretraining with a cross-modality objective; second, supervised fine-tuning for binary PRL classification.

\subsection{Proposed architecture details}
\textit{Two streams backbone.} 
Given paired patches $x^{(Q)}$ (QSM) and $x^{(F)}$ (FLAIR), the network comprises two separate convolutional encoders built using 3D pre-activation residual blocks. Each encoder begins with a $3\times3\times3$ convolution, followed by two \texttt{ResidualBlock3D} modules with stride $1$, a downsampling \\ \texttt{ResidualBlock3D} with stride $(2,2,2)$, and a final \texttt{ResidualBlock3D} with stride $1$.
Each residual block follows a pre-activation design consisting of Group Normalization and ReLU, followed by a $3\times3\times3$ convolution, a second Group Normalization and ReLU, another convolution, and a skip connection. 

\vspace{0.1cm}
\noindent \textit{Spatial FiLM conditioning.}
After two-separate stream encoders, features extracted pass through the \texttt{Spatial FiLM layer}. Let $z^{(Q)}$ and $z^{(F)}$ denote intermediate QSM and FLAIR feature maps. Spatially varying modulation fields $\gamma$ and $\beta$ are generated from $z^{(F)}$ via $1\times1\times1$ convolutions. The modulated QSM representation is:
\begin{equation}
\widetilde{z}^{(Q)} = (1+\gamma)\odot z^{(Q)} + \beta,
\end{equation}
where $\odot$ denotes element-wise multiplication. 
This spatial conditioning enables localized contextual adaptation while preserving QSM as the dominant modality. A $1\times1\times1$ convolution follows feature concatenation before deeper processing.

\vspace{0.1cm}
\noindent \textit{Channel recalibration and classification.}
Squeeze-and-excitation (SE) blocks \cite{hu2018squeeze} are applied at deeper stages to enhance channel selectivity, as shown in Figure \ref{fig:architecture}. The \texttt{SE Block} first squeezes spatial dimensions via 3D Global Average Pooling (GAP), then computes channel-wise weights through a 1x1x1 convolutional bottleneck with reduction and ReLU; finally scales the input tensor element-wisely using Sigmoid activation. After the second \texttt{SE Block}, the final feature map is aggregated via global average pooling and passed to a \texttt{Classification Head} composed by two fully connected layers with ReLU activation and dropout. A sigmoid activation produces the probability of a lesion being Rim$^+$. 

\subsection{Training Strategy}
\textit{Self-supervised multimodal pretraining.}
To improve representation robustness under limited data availability and severe class imbalance, the backbone is first initialized through a self-supervised multimodal pretraining phase \cite{zong2024self,taleb20203d}.
Pretraining is based on cross-modal alignment between intermediate QSM and FLAIR embeddings. Let $z^{(Q)}$ and $z^{(F)}$ denote modality-specific feature representations extracted at the bottleneck stage. We enforce modality-consistent representations through a regression objective $\mathcal{L}_{\mathrm{SSL}} = \| \phi(z^{(F)}) - z^{(Q)} \|_2^2$, where $\phi(\cdot)$ is a lightweight multilayer perceptron mapping QSM features to the FLAIR embedding space. We minimize standard MSE loss between cross-modal embeddings, treating the QSM representation as a target, and enforce that the QSM features are predictable from the corresponding FLAIR features computed on the same spatially transformed patch. During pretraining, random 3D spatial augmentations (axis flips and 90° rotations) are applied consistently across modalities improving robustness \cite{cardoso2022monai}.
After pretraining, the projection head is discarded, the pretrained weights are loaded for the backbone encoders up to the FiLM layers and the model is fine-tuned end-to-end for supervised classification.

\vspace{0.1cm}
\noindent \textit{Supervised fine-tuning.}
During supervised training, the model is optimized using a composite objective. 
For lesion classification, we employ binary cross-entropy (BCE) as the primary loss. 
To improve class separability under severe imbalance, we additionally incorporate a supervised contrastive loss $\mathcal{L}_{\mathrm{contrastive}}$ applied to the global lesion embedding \cite{khosla2020supervised}.
The contrastive term follows the normalized temperature-scaled cross-entropy (NT-Xent) formulation \cite{chen2020simple,mildenberger2025tale}, promoting intra-class compactness and inter-class separation in the embedding space.\\
\noindent The overall training objective is defined as
$
\mathcal{L}_{Sup} = \mathcal{L}_{\mathrm{BCE}} + \lambda \mathcal{L}_{\mathrm{contrastive}},
$
where $\lambda$ controls the relative contribution of the contrastive component.
The model outputs a probability $p \in [0,1]$ representing the likelihood of a lesion being Rim$^+$. A decision threshold $t^\star$ was used to obtain binary predictions: lesions with $p \geq t^\star$ were classified as Rim$^+$, whereas lesions with $p < t^\star$ were classified as Rim$^-$.

\section{Experiments}
\subsection{Dataset and Preprocessing}
The dataset comprises MRI scans from 88 people with MS 
acquired at San Martino Hospital (Genova, Italy) 
using a 3T MAGNETOM Prisma scanner (Siemens). The protocol included T1-weighted (T1w), FLAIR, and QSM sequences. 
Lesion masks were manually segmented and lesion status (Rim$^+$/Rim$^-$) was independently assigned by two expert clinicians. In cases of disagreement,  lesions were jointly reviewed to reach a  consensus. The study was conducted in accordance with the Declaration of Helsinki and approved by the local Institutional Ethics Committee. 
Written informed consent was obtained from all participants.

QSM volumes were acquired at $0.65\times0.65\times0.65$\,mm resolution, while FLAIR volumes were acquired at $1.0\times1.0\times1.0$\,mm resolution. All modalities were rigidly registered to subject-specific T1w space ($1.0\times1.0\times1.0$\,mm resolution) using ANTs \cite{tustison_antsx_2021}. N4 bias field correction was performed to mitigate low-frequency intensity inhomogeneities. Lesion-centered 3D patches were then extracted from instance-level masks, where each connected component represented an individual lesion. Consistent with \cite{tazza2024multiparametric} and clinical expertise, lesions smaller than 110 voxels ($\approx 0.1,mL$) were excluded to ensure reliable morphological characterization.
After preprocessing and size-based exclusion, the final dataset consisted of 1247 lesion-centered patches, of which 90 (7.22\%) were RIM$^+$, reflecting their low prevalence as reported in the literature \cite{tazza2024multiparametric}.
For each lesion $\ell$, the center of mass of its binary mask $M_\ell$ was used to extract a fixed $64\times64\times64$ voxel patch from both modalities, using zero-padding when necessary. Intensity normalization was performed independently per modality: FLAIR volumes were scaled using robust percentile normalization (0.5–99.5\%) per participant, while QSM volumes were clipped to $[-300,300]$ and linearly rescaled to preserve susceptibility contrast.
To constrain spatial support to lesion-centered context, we applied adaptive 3D morphological dilation to the ground-truth mask and suppressed all voxels outside the dilated region. The dilation radius was derived from lesion volume (scaled by $k=0.75$) 
. The resulting mask was used to gate the patch, suppressing unrelated background tissue while preserving perilesional information.

\subsection{Experimental Setup}
We adopted a five-fold evaluation protocol at the participant level to prevent lesion-level data leakage and assess generalizability across different people. Participants were manually stratified to ensure balanced distribution of total lesion burden and number of RIM-positive individuals across folds. Each fold contained 16–18 participants, including 6–8 RIM-positive participants.
For each evaluation run, one fold was held out as an independent test set, while the remaining four folds constituted the development set. Within the development set, lesions were randomly split into training (80\%) and validation (20\%) subsets. The training subset was used for model optimization, while the validation subset was used for early stopping and threshold $t^\star$ selection. 
This procedure was repeated five times, such that each participant served as test data exactly once. Final performance was computed by aggregating results across the five test folds.
To address class imbalance, a WeightedRandomSampler was used to generate approximately class-balanced mini-batches by sampling lesions with probability inversely proportional to their class frequency.

\vspace{0.1cm}
\noindent \textit{Implementation Details}
All experiments were implemented in PyTorch and executed on an NVIDIA H200 NVL GPU (MIG-enabled, up to 35 GB memory) using CUDA 13.0. Mixed-precision (bfloat16) training was employed. Pretraining is performed using AdamW \cite{loshchilov2017decoupled} (learning rate $5\times10^{-5}$, weight decay $0.05$) with batch size 10, mixed precision (bfloat16), early stopping (patience 20). 
During supervised training, the balancing coefficient $\lambda$ is learned via a non-negative parametrization and the temperature parameter was set to $\tau = 0.1$.
Optimization was performed using AdamW with gradient clipping. Early stopping was driven by validation loss during self-supervised pretraining, and by validation PR-AUC during  supervised fine-tuning.

\vspace{0.1cm}
\noindent \textit{Evaluation}
Performance was assessed at the lesion level under the fixed 5-fold evaluation protocol. Primary threshold-independent evaluation metrics were Precision–Recall AUC (PR-AUC)and ROC-AUC; for each fold, lesion-wise probabilities were generated on held-out test participants, and these metrics were computed from the full score distributions. Threshold-dependent metrics such as accuracy, sensitivity, specificity, positive predictive value (PPV) and F1-score were computed using an optimized threshold $t^\star$ selected on the validation split by maximizing the F1-score via grid search over $t\in[0,1]$. 
Summary metrics were obtained by averaging fold-wise metrics (mean ± standard deviation).
The same evaluation protocol and data splits were applied to all compared models.

\subsection{Results}

\begin{figure}[htbp]
    \centering
    \includegraphics[width=0.49\linewidth]{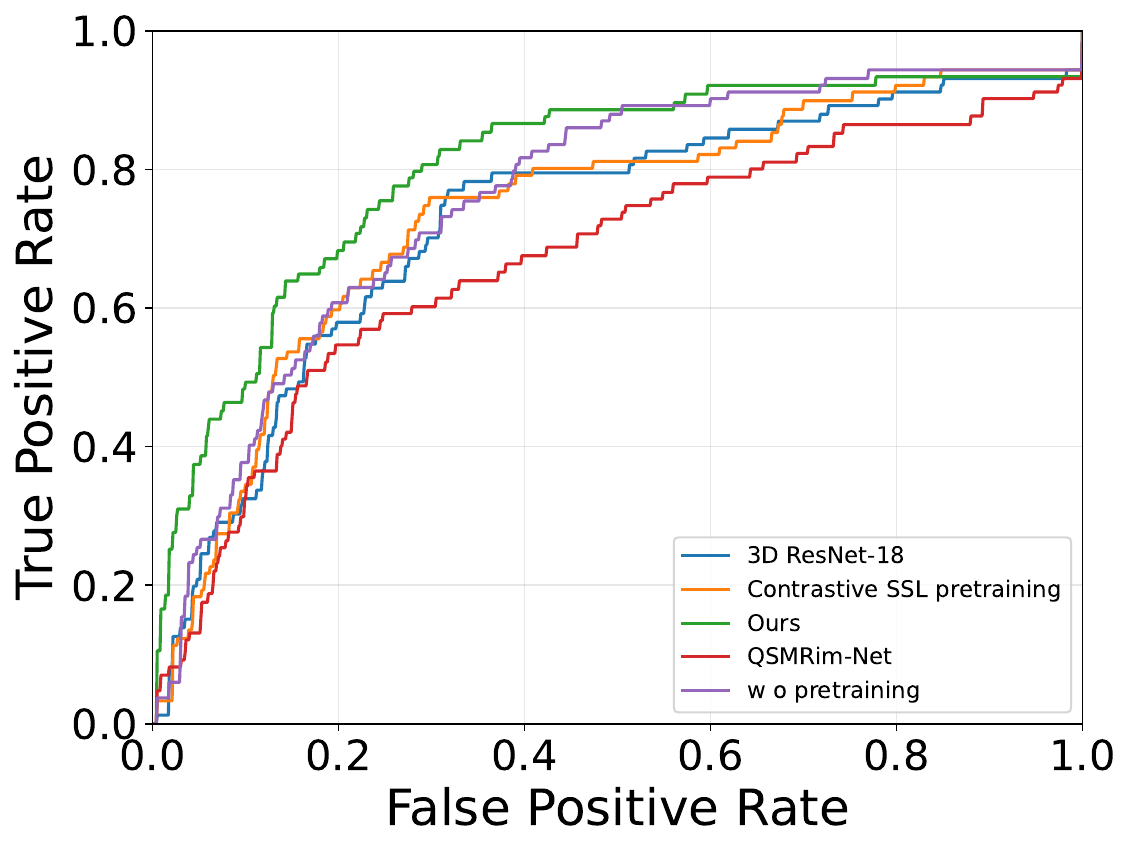}
    \hfill
    \includegraphics[width=0.49\linewidth]{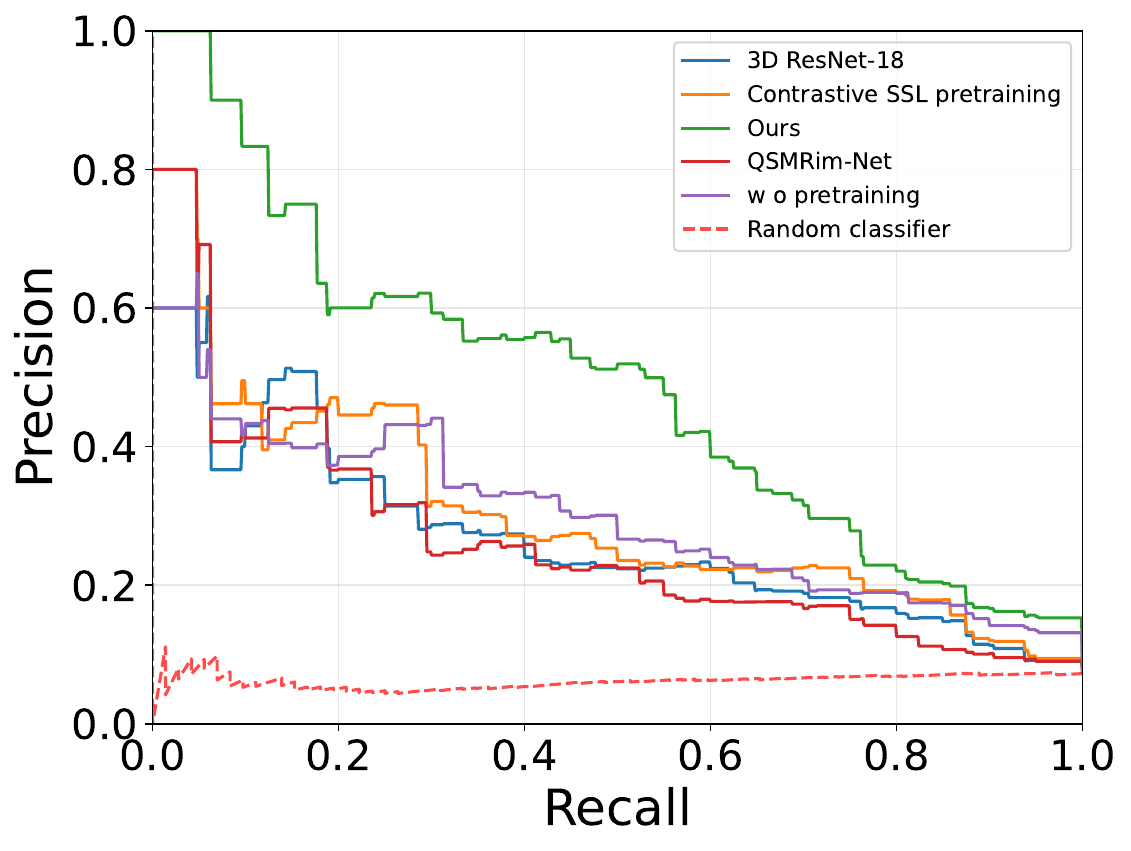}
    \caption{Lesion-wise ROC and PR curves for the five models. The displayed are 5-fold evaluation protocol mean curves, obtaining by interpolating each fold onto a common FPR grid and averaging TPR point-wise. The PR curve also reports a simulated random-classifier baseline with positive-class prevalence of $7.22\%$. No confidence bands are reported for readability.}
    \label{fig:roc_pr}
\end{figure}

We evaluate the proposed architecture against QSMRim-Net~\cite{zhang2022qsmrim} and a 3D ResNet-18 implemented in MONAI~\cite{he2016deep,cardoso2022monai} for binary classification. In addition, we conduct ablation studies to isolate the contribution of the contrastive component and the pretraining strategy. Specifically, we report: (i) training from scratch using the composite objective $\mathcal{L}_{\mathrm{BCE}} + \lambda \mathcal{L}_{\mathrm{contrastive}}$; and (ii) self-supervised contrastive pretraining with NT-Xent objective \cite{chaitanya2020contrastive} followed by supervised fine-tuning with $\mathcal{L}_{\mathrm{BCE}}$. All models were trained using a WeightedRandomSampler to account for class imbalance. QSMRim-Net \cite{zhang2022qsmrim} was excluded from this setting, as it incorporates DeepSMOTE within its framework to handle imbalance. We therefore evaluated QSMRim-Net re-training the model and tested on our dataset with the same protocol evaluation, using the authors’ official GitHub implementation. During SSL contrastive pretraining, the temperature parameter was set to $\tau_{\mathrm{SSL}} = 0.1$.

\begin{table*}[ht]
\centering
\caption{Area-under-curve metrics at the lesion level (averaged across AUCs test folds). Values are reported as mean $\pm$ standard deviation.}
\label{tab:auc_results}
\renewcommand{\arraystretch}{1.15}

\begin{tabularx}{\textwidth}{|l|
>{\centering\arraybackslash}X
>{\centering\arraybackslash}X|}
\hline
\textbf{Method} 
& \textbf{ROC AUC $\uparrow$ } 
& \textbf{PR AUC $\uparrow$ }
\\
\hline

QSMRim-Net \cite{zhang2022qsmrim}
& 0.658 $\pm$ 0.091& 0.156 $\pm$ 0.100 \\

3D ResNet-18 \cite{cardoso2022monai}
& 0.726 $\pm$ 0.052 & 0.234 $\pm$ 0.026\\
\hline
w/o pretraining + $\mathcal{L}_{\mathrm{contrastive}}$
& 0.789 $\pm$  0.067 & 0.271 $\pm$ 0.125\\

Contrastive SSL pretraining
& 0.800 $\pm$ 0.031 &  0.256 $\pm$ 0.129\\
\hline
\textbf{Ours}
&\textbf{ 0.869 $\pm$ 0.049} & \textbf{0.457 $\pm$ 0.141}  \\

\hline
\end{tabularx}

\end{table*}

The proposed model achieves the highest discriminative performance across all metrics, see Fig. \ref{fig:roc_pr} and Table \ref{tab:auc_results}, with ROC AUC $0.869 \pm 0.049$ and PR AUC $0.457 \pm 0.141$, substantially outperforming both QSMRim-Net and the ResNet baseline. Under strong class imbalance (7.22\% RIM prevalence), PR AUC provides a more faithful estimate of minority-class discrimination. The improvement in PR AUC indicates that the learned representation better separates positive from negative samples in regions of low recall, suggesting improved feature compactness and class-conditional separability. Table \ref{tab:lesion_results} reports the performances obtained by the optimized threshold. The benefit of our method is primarily driven by increased precision (PPV $0.437 \pm 0.099$) while maintaining competitive sensitivity ($0.447 \pm 0.221$) and high specificity ($0.956 \pm 0.021$), indicating a reduction in false-positive activations without sacrificing recall. The higher F1-score ($0.460 \pm 0.155$) further confirms a more balanced precision–recall trade-off.
These results suggest that multimodal conditioning combined with contrastive objectives induces a more structured embedding space. 
Ablation analyses further support this interpretation. Removing SSL pretraining leads to consistent degradation in both ROC AUC and PR AUC, with a particularly pronounced drop in PR performance. This behavior indicates that cross-modal pretraining enhances minority-class feature alignment prior to supervised optimization. Overall, the combination of spatial conditioning and contrastive pretraining yields a more discriminative and stable lesion-level representation compared to purely supervised baselines.

\begin{table*}[htbp]
\centering
\caption{Lesion-level performance comparison across test folds. Values are reported as mean $\pm$ standard deviation.}
\label{tab:lesion_results}
\renewcommand{\arraystretch}{1.15}

\begin{tabularx}{\textwidth}{|l|
>{\centering\arraybackslash}X
>{\centering\arraybackslash}X
>{\centering\arraybackslash}X|}
\hline
\textbf{Method} 
& \textbf{Accuracy}
& \textbf{F1-score}
& \textbf{Sensitivity}
\\
\hline

QSMRim-Net \cite{zhang2022qsmrim}
& 0.855 $\pm$ 0.058 & 0.235 $\pm$ 0.072  & 0.338 $\pm$ 0.180 \\

3D ResNet-18 \cite{cardoso2022monai}
& 0.898 $\pm$ 0.028 & 0.211 $\pm$ 0.122 & 0.189 $\pm$ 0.136 \\
\hline
w/o pretraining + $\mathcal{L}_{\mathrm{contrastive}}$
& 0.864 $\pm$ 0.077 & 0.325 $\pm$ 0.063 & 0.444 $\pm$ 0.107 \\

Contrastive SSL pretraining
& 0.877 $\pm$ 0.056 & 0.319 $\pm$ 0.081 & 0.400 $\pm$0.128  \\
\hline
\textbf{Ours}
& \textbf{0.921 $\pm$ 0.012} 
& \textbf{0.460 $\pm$ 0.155} 
& \textbf{0.447 $\pm$ 0.221} \\

\hline
\end{tabularx}

\vspace{0.4cm}

\begin{tabularx}{\textwidth}{|l|
>{\centering\arraybackslash}X
>{\centering\arraybackslash}X|}
\hline
\textbf{Method} 
& \textbf{Specificity}
& \textbf{PPV}
\\
\hline

QSMRim-Net \cite{zhang2022qsmrim}
& 0.893 $\pm$ 0.072 & 0.220 $\pm$  0.063 \\

3D ResNet-18 \cite{cardoso2022monai}
& 0.946 $\pm$ 0.035 & 0.239 $\pm$ 0.184 \\
\hline
w/o pretraining + $\mathcal{L}_{\mathrm{contrastive}}$
& 0.895 $\pm$ 0.085 & 0.253 $\pm$ 0.095 \\

Contrastive SSL pretraining
& 0.914 $\pm$ 0.066 & 0.265 $\pm$ 0.129 \\
\hline
\textbf{Ours}
& \textbf{0.956 $\pm$ 0.021} 
& \textbf{0.437 $\pm$ 0.099} \\
\hline
\end{tabularx}
\end{table*}

\noindent DeLong’s test \cite{delong1988comparing} for correlated ROC curves was applied to out-of-fold lesion-level predictions, concatenated over the 5 test sets. 
The proposed model significantly outperformed QSMRimNet in terms of ROC-AUC 
(0.839 vs 0.658; $\Delta=0.181$; $p-value<0.001$), 
with improvements observed across all test folds.

\section{Conclusion}
We presented an end-to-end multimodal framework for lesion-level classification of PRL in Multiple Sclerosis using QSM and FLAIR MRI. The proposed architecture models modality asymmetry by prioritizing susceptibility-driven information from QSM while structurally conditioning it with complementary FLAIR context. Evaluation on a clinically representative cohort highlight improved lesion-level discrimination compared to unimodal and baseline architectures, supporting robust automated Rim$^+$/Rim$^-$ classification without reliance on handcrafted radiomic features. Future work will focus on validation across multi-center datasets, robustness to acquisition variability, integration within end-to-end lesion detection and longitudinal monitoring pipelines.

\bibliographystyle{splncs04}
\bibliography{refs}

@article{delong1988comparing,
  title={Comparing the areas under two or more correlated receiver operating characteristic curves: a nonparametric approach},
  author={DeLong, Elizabeth R and DeLong, David M and Clarke-Pearson, Daniel L},
  journal={Biometrics},
  pages={837--845},
  year={1988},
  publisher={JSTOR}
}

@article{chawla2002smote,
  title={SMOTE: synthetic minority over-sampling technique},
  author={Chawla, Nitesh V and Bowyer, Kevin W and Hall, Lawrence O and Kegelmeyer, W Philip},
  journal={Journal of artificial intelligence research},
  volume={16},
  pages={321--357},
  year={2002}
}

@inproceedings{he2016deep,
  title={Deep residual learning for image recognition},
  author={He, Kaiming and Zhang, Xiangyu and Ren, Shaoqing and Sun, Jian},
  booktitle={Proceedings of the IEEE conference on computer vision and pattern recognition},
  pages={770--778},
  year={2016}
}

@article{loshchilov2017decoupled,
  title={Decoupled weight decay regularization},
  author={Loshchilov, Ilya and Hutter, Frank},
  journal={arXiv preprint arXiv:1711.05101},
  year={2017}
}

@inproceedings{perez2018film,
  title={Film: Visual reasoning with a general conditioning layer},
  author={Perez, Ethan and Strub, Florian and De Vries, Harm and Dumoulin, Vincent and Courville, Aaron},
  booktitle={Proceedings of the AAAI conference on artificial intelligence},
  volume={32},
  number={1},
  year={2018}
}

@inproceedings{hu2018squeeze,
  title={Squeeze-and-excitation networks},
  author={Hu, Jie and Shen, Li and Sun, Gang},
  booktitle={Proceedings of the IEEE conference on computer vision and pattern recognition},
  pages={7132--7141},
  year={2018}
}

@article{barquero2020rimnet,
  title={RimNet: A deep 3D multimodal MRI architecture for paramagnetic rim lesion assessment in multiple sclerosis},
  author={Barquero, Germ{\'a}n and La Rosa, Francesco and Kebiri, Hamza and Lu, Po-Jui and Rahmanzadeh, Reza and Weigel, Matthias and Fartaria, M{\'a}rio Jo{\~a}o and Kober, Tobias and Th{\'e}audin, Marie and Du Pasquier, Renaud and others},
  journal={NeuroImage: Clinical},
  volume={28},
  pages={102412},
  year={2020},
  publisher={Elsevier}
}

@inproceedings{chen2020simple,
  title={A simple framework for contrastive learning of visual representations},
  author={Chen, Ting and Kornblith, Simon and Norouzi, Mohammad and Hinton, Geoffrey},
  booktitle={International conference on machine learning},
  pages={1597--1607},
  year={2020},
  organization={PmLR}
}

@article{taleb20203d,
  title={3d self-supervised methods for medical imaging},
  author={Taleb, Aiham and Loetzsch, Winfried and Danz, Noel and Severin, Julius and Gaertner, Thomas and Bergner, Benjamin and Lippert, Christoph},
  journal={Advances in neural information processing systems},
  volume={33},
  pages={18158--18172},
  year={2020}
}

@article{khosla2020supervised,
  title={Supervised contrastive learning},
  author={Khosla, Prannay and Teterwak, Piotr and Wang, Chen and Sarna, Aaron and Tian, Yonglong and Isola, Phillip and Maschinot, Aaron and Liu, Ce and Krishnan, Dilip},
  journal={Advances in neural information processing systems},
  volume={33},
  pages={18661--18673},
  year={2020}
}

@article{chaitanya2020contrastive,
  title={Contrastive learning of global and local features for medical image segmentation with limited annotations},
  author={Chaitanya, Krishna and Erdil, Ertunc and Karani, Neerav and Konukoglu, Ender},
  journal={Advances in neural information processing systems},
  volume={33},
  pages={12546--12558},
  year={2020}
}

@article{tustison_antsx_2021,
	title = {The {ANTsX} ecosystem for quantitative biological and medical imaging},
	volume = {11},
	issn = {2045-2322},
	url = {https://doi.org/10.1038/s41598-021-87564-6},
	doi = {10.1038/s41598-021-87564-6},
	number = {1},
	journal = {Scientific Reports},
	author = {Tustison, Nicholas J. and Cook, Philip A. and Holbrook, Andrew J. and Johnson, Hans J. and Muschelli, John and Devenyi, Gabriel A. and Duda, Jeffrey T. and Das, Sandhitsu R. and Cullen, Nicholas C. and Gillen, Daniel L. and Yassa, Michael A. and Stone, James R. and Gee, James C. and Avants, Brian B.},
	month = apr,
	year = {2021},
	pages = {9068},
}

@article{lou2021fully,
  title={Fully automated detection of paramagnetic rims in multiple sclerosis lesions on 3T susceptibility-based MR imaging},
  author={Lou, Carolyn and Sati, Pascal and Absinta, Martina and Clark, Kelly and Dworkin, Jordan D and Valcarcel, Alessandra M and Schindler, Matthew K and Reich, Daniel S and Sweeney, Elizabeth M and Shinohara, Russell T},
  journal={NeuroImage: Clinical},
  volume={32},
  pages={102796},
  year={2021},
  publisher={Elsevier}
}

@article{tranfa2022quantitative,
  title={Quantitative MRI in multiple sclerosis: from theory to application},
  author={Tranfa, M and Pontillo, G and Petracca, M and Brunetti, A and Tedeschi, E and Palma, G and Cocozza, S},
  journal={American Journal of Neuroradiology},
  volume={43},
  number={12},
  pages={1688--1695},
  year={2022},
  publisher={American Journal of Neuroradiology}
}

@article{zhang2022qsmrim,
  title={QSMRim-Net: Imbalance-aware learning for identification of chronic active multiple sclerosis lesions on quantitative susceptibility maps},
  author={Zhang, Hang and Nguyen, Thanh D and Zhang, Jinwei and Marcille, Melanie and Spincemaille, Pascal and Wang, Yi and Gauthier, Susan A and Sweeney, Elizabeth M},
  journal={NeuroImage: Clinical},
  volume={34},
  pages={102979},
  year={2022},
  publisher={Elsevier}
}

@article{cardoso2022monai,
  title={Monai: An open-source framework for deep learning in healthcare},
  author={Cardoso, M Jorge and Li, Wenqi and Brown, Richard and Ma, Nic and Kerfoot, Eric and Wang, Yiheng and Murrey, Benjamin and Myronenko, Andriy and Zhao, Can and Yang, Dong and others},
  journal={arXiv preprint arXiv:2211.02701},
  year={2022}
}

@article{dal2024chronic,
  title={Chronic active lesions in multiple sclerosis: classification, terminology, and clinical significance},
  author={Dal-Bianco, Assunta and Oh, Jiwon and Sati, Pascal and Absinta, Martina},
  journal={Therapeutic Advances in Neurological Disorders},
  volume={17},
  pages={17562864241306684},
  year={2024},
  publisher={SAGE Publications Sage UK: London, England}
}

@article{zong2024self,
  title={Self-supervised multimodal learning: A survey},
  author={Zong, Yongshuo and Mac Aodha, Oisin and Hospedales, Timothy M},
  journal={IEEE Transactions on Pattern Analysis and Machine Intelligence},
  volume={47},
  number={7},
  pages={5299--5318},
  year={2024},
  publisher={IEEE}
}

@inproceedings{mildenberger2025tale,
  title={A tale of two classes: adapting supervised contrastive learning to binary imbalanced datasets},
  author={Mildenberger, David and Hager, Paul and Rueckert, Daniel and Menten, Martin J},
  booktitle={Proceedings of the Computer Vision and Pattern Recognition Conference},
  pages={10305--10314},
  year={2025}
}

@article{kuhlmann2017updated,
  title={An updated histological classification system for multiple sclerosis lesions},
  author={Kuhlmann, Tanja and Ludwin, Samuel and Prat, Alexandre and Antel, Jack and Br{\"u}ck, Wolfgang and Lassmann, Hans},
  journal={Acta neuropathologica},
  volume={133},
  number={1},
  pages={13--24},
  year={2017},
  publisher={Springer}
}

@article{absinta2019association,
  title={Association of chronic active multiple sclerosis lesions with disability in vivo},
  author={Absinta, Martina and Sati, Pascal and Masuzzo, Federica and Nair, Govind and Sethi, Varun and Kolb, Hadar and Ohayon, Joan and Wu, Tianxia and Cortese, Irene CM and Reich, Daniel S},
  journal={JAMA neurology},
  volume={76},
  number={12},
  pages={1474--1483},
  year={2019}
}

@article{elliott2023lesion,
  title={Lesion-level correspondence and longitudinal properties of paramagnetic rim and slowly expanding lesions in multiple sclerosis},
  author={Elliott, Colm and Rudko, David A and Arnold, Douglas L and Fetco, Dumitru and Elkady, Ahmed M and Araujo, David and Zhu, Bing and Gafson, Arie and Tian, Zhe and Belachew, Shibeshih and others},
  journal={Multiple Sclerosis Journal},
  volume={29},
  number={6},
  pages={680--690},
  year={2023},
  publisher={SAGE Publications Sage UK: London, England}
}

@article{montalban2025diagnosis,
  title={Diagnosis of multiple sclerosis: 2024 revisions of the McDonald criteria},
  author={Montalban, Xavier and Lebrun-Fr{\'e}nay, Christine and Oh, Jiwon and Arrambide, Georgina and Moccia, Marcello and Amato, Maria Pia and Amezcua, Lilyana and Banwell, Brenda and Bar-Or, Amit and Barkhof, Frederik and others},
  journal={The Lancet Neurology},
  volume={24},
  number={10},
  pages={850--865},
  year={2025},
  publisher={Elsevier}
}

@article{rahmanzadeh2022new,
  title={A new advanced MRI biomarker for remyelinated lesions in multiple sclerosis},
  author={Rahmanzadeh, Reza and Galbusera, Riccardo and Lu, Po-Jui and Bahn, Erik and Weigel, Matthias and Barakovic, Muhamed and Franz, Jonas and Nguyen, Thanh D and Spincemaille, Pascal and Schiavi, Simona and others},
  journal={Annals of neurology},
  volume={92},
  number={3},
  pages={486--502},
  year={2022},
  publisher={Wiley Online Library}
}

@article{tazza2024multiparametric,
  title={Multiparametric characterization and spatial distribution of different MS lesion phenotypes},
  author={Tazza, Francesco and Boffa, Giacomo and Schiavi, Simona and Lapucci, Caterina and Piredda, Gian Franco and Cipriano, Emilio and Zac{\`a}, Domenico and Roccatagliata, Luca and Hilbert, Tom and Kober, Tobias and others},
  journal={American Journal of Neuroradiology},
  volume={45},
  number={8},
  pages={1166--1174},
  year={2024},
  publisher={American Journal of Neuroradiology}
}
\end{document}